\documentclass[
]{ceurart}

\usepackage{listings}
\usepackage{hyperref}

\usepackage{comment}
\usepackage{adjustbox}
\usepackage{multirow}
\usepackage{booktabs}
\usepackage{csquotes}
\usepackage{amsmath}
\usepackage{graphicx}
\graphicspath{ {./images/} }
\usepackage{svg}
\usepackage{threeparttable}
\begin{document}

\copyrightyear{2024}
\copyrightclause{Copyright for this paper by its authors.
  Use permitted under Creative Commons License Attribution 4.0
  International (CC BY 4.0).}

\conference{CLEF 2024: Conference and Labs of the Evaluation Forum, September 09–12, 2024, Grenoble, France}

\title{Exploring the Latest LLMs for Leaderboard Extraction}

\title[mode=sub]{Notebook for the SimpleText Task4: SOTA? Lab at CLEF 2024}

\author[1]{Salomon Kabongo}[%
orcid=0000-0002-0021-9729,
email=kabenamualu@l3s.de,
url=https://skabongo.github.io/,
]
\cormark[1]
\fnmark[1]
\address[1]{Leibniz University of Hannover, Hannover, Germany}

\author[1,2]{Jennifer D'Souza}[%
orcid=0000-0002-6616-9509,
email=jennifer.dsouza@tib.eu,
% url=https://kmitd.github.io/ilaria/,
]
\fnmark[1]
\address[2]{TIB Leibniz Information Centre for Science and Technology, Hannover, Germany}
\author[1,2]{S\"oren Auer}[%
orcid=0000-0002-0698-28646,
email=auer@tib.eu,
% url=http://conceptbase.sourceforge.net/mjf/,
]
% \address[4]{University of Skövde, Högskolevägen 1, 541 28 Skövde, Sweden}

% \author{Salomon Kabongo\inst{1}\orcidID{0000-0002-0021-9729} \and
% Jennifer D'Souza\inst{2,3}\orcidID{0000-0002-6616-9509} \and
% S\"oren Auer\inst{3}\orcidID{0000-0002-0698-2864}}

% \authorrunning{S. Kabongo et al.}
% % \authorrunning{anonymous}
% \institute{L3S Research Center, Leibniz University of Hannover, Hannover, Germany
% \email{kabenamualu@l3s.de}\\ \and
% TIB Leibniz Information Centre for Science and Technology, Hannover, Germany\\
% \email{\{jennifer.dsouza,auer\}@tib.eu}}

\cortext[1]{Corresponding author.}
\fntext[1]{These authors contributed equally.}

\begin{abstract}
    The rapid advancements in Large Language Models (LLMs) have opened new avenues for automating complex tasks in AI research. This paper investigates the efficacy of different LLMs-Mistral 7B, Llama-2, GPT-4-Turbo and GPT-4.o in extracting leaderboard information from empirical AI research articles. We explore three types of contextual inputs to the models: DocTAET (Document Title, Abstract, Experimental Setup, and Tabular Information), DocREC (Results, Experiments, and Conclusions), and DocFULL (entire document). Our comprehensive study evaluates the performance of these models in generating (Task, Dataset, Metric, Score) quadruples from research papers. The findings reveal significant insights into the strengths and limitations of each model and context type, providing valuable guidance for future AI research automation efforts.
\end{abstract}


\maketitle
\begin{abstract}
This paper explores the impact of context selection on the efficiency of Large Language Models (LLMs) in generating Artificial Intelligence (AI) research leaderboards, a task defined as the extraction of (Task, Dataset, Metric, Score) quadruples from scholarly articles. By framing this challenge as a text generation objective and employing instruction finetuning with the FLAN-T5 collection, we introduce a novel method that surpasses traditional Natural Language Inference (NLI) approaches in adapting to new developments without a predefined taxonomy. Through experimentation with three distinct context types of varying selectivity and length, our study demonstrates the importance of effective context selection in enhancing LLM accuracy and reducing hallucinations, providing a new pathway for the reliable and efficient generation of AI leaderboards. This contribution not only advances the state of the art in leaderboard generation but also sheds light on strategies to mitigate common challenges in LLM-based information extraction.
\end{abstract}

\section{Introduction}
% Mention sota in the and share task in the introduction 
% Link to the simple text initiative 

The unprecedented expansion of scientific publications~\cite{fortunato2018science,bornmann2021growth}  in the field of artificial intelligence (AI) has created a substantial challenge in systematically tracking and assessing advancements. Leaderboards, which rank AI models based on their performance across various tasks and datasets, serve as a crucial mechanism for monitoring progress and driving competition. The extraction of state-of-the-art information, represented by (Task, Dataset, Metric, Score) quadruples, is vital for the maintenance and accuracy of these leaderboards.

The effectiveness of LLMs in information extraction (IE) tasks is deeply influenced by the context provided during the input phase. In this context, “context” refers to the segments of the scholarly articles from which the (Task, Dataset, Metric, Score) data is extracted. Unlike the concept of in-context learning, where models are primed with examples, our focus is on selecting relevant sections of text to guide the LLMs toward accurate and relevant information extraction while minimizing the risk of generating irrelevant or false information (hallucinations).

\begin{figure}[!tb]
\includegraphics[width=\linewidth]{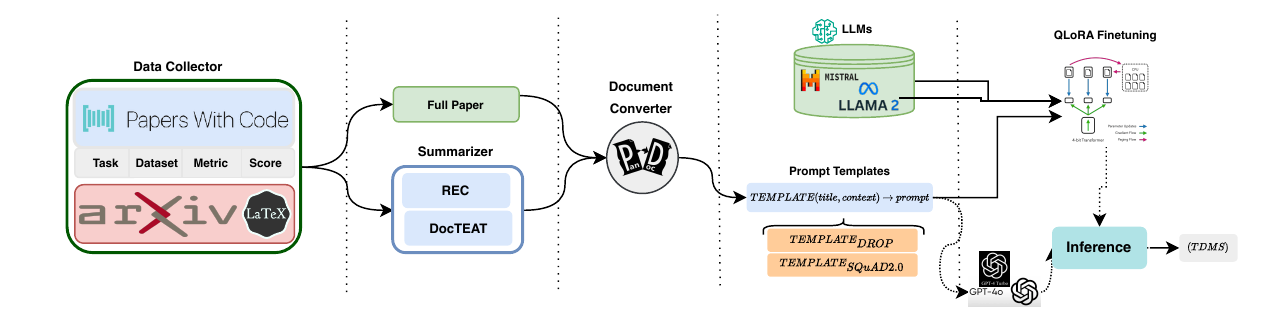}
\caption{Main process overview}
\label{fig:main_process}
\end{figure}

Existing research has shown that longer contextual inputs can lead to decreased model accuracy and higher rates of hallucination due to the potential distraction of irrelevant information~\cite{shi2023large,Liu2023Lost}. Therefore, identifying the optimal length and specificity of context is crucial~\cite{shi2023large}. This study explores three distinct types of contextual inputs to empirically evaluate their impact on the task of extracting leaderboards: DocTAET (Document Title, Abstract, Experimental Setup, and Tabular Information), DocREC (Results, Experiments, and Conclusions), and DocFULL (the entire document).

% To contribute to this effort, 

Our primary contributions are twofold. Firstly, we evaluate the performance of the latest LLMs, both open-sourced Llama-2 and Mistral 7B~\cite{touvron2023llama,mistral_v01} and proprietary GPT-4-Turbo and GPT-4.o~\cite{achiam2023gpt}, in generating structured summaries and classifying papers with and without leaderboard data. Our first research question (RQ1) encapsulates this: Which LLM provides the most accurate performance for generating structured summaries and for leaderboard/no-leaderboard classification?

Secondly, we focus on the precise extraction of individual elements, addressing our second research question (RQ2): Which LLM offers the best trade-off between precision and other performance metrics in few-shot and zero-shot settings? This question is critical, given the importance of precision in scholarly communications and the broader implications for the reliability of model outputs.

Additionally, we participate in the \href{https://sites.google.com/view/simpletext-sota/home}{\textit{“SOTA? Tracking the State-of-the-Art in Scholarly Publications”}} shared Task 4 in the SimpleText\footnote{https://simpletext-project.com/2024/en/} track of CLEF 2024 \cite{sotaTrackCLEF2024,simpleTextCLEF2024}. The goal of the SOTA? shared task is to develop systems that, given the full text of an AI paper, can recognize whether the paper reports model scores on benchmark datasets and, if so, extract all pertinent (Task, Dataset, Metric, Score) tuples presented within the paper.

\section{Related Work}

% History 
The aspiration for automatically generated leaderboards to monitor advancements in scientific research has been a notable ambition within the scientific community. This initiative first gained traction through the analysis of citation networks, employing then-cutting-edge methodologies such as Markov Random Fields and others \cite{qazvinian-radev-2010-identifying,athar-teufel-2012-detection}.

The advent of transformer-based models \cite{vaswani2017attention} marked a significant leap forward, setting new benchmarks across a myriad of machine learning tasks. In this vein, researchers at IBM \cite{hou-etal-2019-identification} utilized the Bert pre-trained model \cite{devlin2018bert} within a Natural Language Inference (NLI) framework to discern entailment from the complete texts of research papers to their corresponding leaderboards.

In the realm of information extraction, the application of LLMs such as GPT-3.5 and LLAMA has demonstrated significant promise. A study by \cite{hu2024improving} highlighted the capabilities of GPT-3.5 and GPT-4 in clinical named entity recognition (NER), showcasing their adeptness at processing intricate clinical datasets with limited prerequisite training. Through strategic prompt engineering, these models exhibited remarkable improvements in performance for extracting medical entities from clinical documentation, reinforcing the potential of LLMs in executing complex NER tasks within the healthcare sector \cite{hu2024improving}.

The emergence of LLMs, including ChatGPT, has sparked a reconsideration of specialized versus general-purpose training approaches in the context of LLMs. Building upon the foundation of utilizing LLMs for specialized information extraction tasks, \cite{shamsabadi-etal-2024-large} delves into the utilization of these models in the domain of virology. The study showcases how LLMs, specifically tuned for scientific content, can efficiently parse and extract virology-related information from a plethora of scientific publications. This research underlines the importance of fine-tuning and prompt engineering in enhancing the model's ability to discern relevant scientific facts, contributing to the development of domain-specific leaderboards. The approach exemplified in this paper demonstrates an effective strategy for context selection in LLM-based leaderboard generation, emphasizing the necessity for domain-specific adjustments to maximize the accuracy and relevance of the extracted information.

In our previous study \cite{kabongo2024context}, we expanded on these findings by empirically investigating different ways to select context in creating leaderboards. We analyze the impact of tailored context cues and the integration of domain-specific knowledge on the precision and utility of automatically generated leaderboards. Our contributions included a thorough assessment of the latest LLMs, both open-source and proprietary, for assessing the effectiveness of different context selection strategies and the development of a novel methodology that significantly enhances the accuracy and relevance of LLM-based leaderboard generation. This study not only corroborates the pivotal role of context in leveraging LLMs for information extraction but also introduces innovative techniques that refine the process of generating insightful and reliable leaderboards in the scientific community.

\begin{table*}[h]
\begin{center}
\begin{threeparttable}
\begin{minipage}{\textwidth}
\begin{tabular*}{\textwidth}{@{\extracolsep{\fill}}l|ccc@{\extracolsep{\fill}}}
% \toprule%
\cmidrule{1-4}%
& \multicolumn{3}{@{}c@{}}{\textbf{Our Corpus}} \\\cmidrule{2-4}%
 & Train &Test-Few-shot & Test Zero-shot \\
\midrule
Papers w/ leaderboards & 7,987 & 753& 241  \\
Papers w/o leaderboards &  4,401 & 648 & 548  \\
Total TDM-triples & 415,788 & 34,799 & 14,800  \\
%Avg. TDMS Tokens & 4.1\tnote{a} & 4.1\tnote{a} & 46.23 & 45.19 &  2.64\tnote{a} & 2.41\tnote{a} \\
Distinct TDM-triples & 11,998 & 1,917 & 1,267  \\
Distinct \textit{Tasks}       & 1,374 & 322 & 236  \\
Distinct \textit{Datasets}    & 4,816 & 947 & 647  \\
Distinct \textit{Metrics}     & 2,876 & 654 & 412  \\
Avg. no. of TDM per paper & 5.12 & 4.81 & 6.11  \\
Avg. no. of TDMS per paper & 6.95 & 5.81 & 7.86\\
% \botrule
\end{tabular*}
\caption{Our DocREC (Documents Result[s], Experimentation[s] and Conclusion) corpora statistics. The ``papers w/o leaderboard'' refers to papers that do not report leaderboard.}
%\begin{tablenotes}
%    \item[a] Avg. number of TDM-triples per paper
%\end{tablenotes}
\label{table:DocREC_Stats}
\end{minipage}
\end{threeparttable}
\end{center}
\end{table*}

\section{Methodology}
\label{sec:methodology}
This section details the methodology employed in our study, encompassing data collection, preprocessing, model selection, and evaluation metrics. The goal is to systematically evaluate the performance of four state-of-the-art LLMs—Mistral 7B, Llama-2, GPT-4-Turbo and GPT-4.o—across different context types for the task of leaderboard extraction.

\subsection{Data Collection and Preprocessing}
\label{sec:dataset}
We utilized data that was previously published by~\cite{kabongo2024context} and re-released as the SOTA? shared task training corpus. The dataset consists of (T, D, M, S) annotations for thousands of AI articles available on PwC (CC BY-SA). These articles cover various AI domains such as Natural Language Processing, Computer Vision, Robotics, Graphs, Reasoning, etc., making them representative for empirical AI research. The specific PwC source was downloaded on December 09, 2023. The corpus comprised over 8,000 articles, with 7,987 used for training and 994 for testing, including 751 in the few-shot setting and 241 in the zero-shot setting. These articles, originally sourced from arXiv under CC-BY licenses, are available as latex source code, each accompanied by one or more (T, D, M, S) annotations from PwC. The articles' metadata was directly obtained from the PwC data release, and the articles collection had to be reconstructed by downloading them from arXiv under CC-BY licenses. 

After downloading the article's source code (`.tex`), we needed to preprocess it to convert it to plain text. Sometimes, articles written in LaTeX are split into multiple files. To address this, we first created and executed a custom script to merge the project source code into a single LaTeX file corresponding to the arXiv ID of the paper. Next, we used another custom script to extract specific sections of the paper (DocTEAT or DocREC) from the $`arxiv\_id.tex`$ file, ensuring that the file remained compilable by LaTeX, which is necessary for our $`tex\_to\_text`$ parsing process.

To convert the resulting all-in-one $`arxiv\_id.tex$` file to plain text, we used the command $`pandoc --to=plain`$. Subsequently, each article's parsed text was annotated with (T, D, M, S) quadruples via distant labeling. The overall corpus statistics are reported in Table \ref{table:DocREC_Stats}.

Another important subset of our data, in addition to our base dataset reported in \autoref{table:DocREC_Stats}, was the "no leaderboards papers". We included a set of approximately 4,401 and 648 articles that do not report leaderboards into the train and test sets, respectively. These articles were randomly selected by leveraging the arxiv category feature, then filtering it to papers belonging to domains unrelated to AI/ML/Stats. These articles were annotated with the \textit{unanswerable} label to finetune our language model in recognizing papers without (T,D,M,S) mentions in them.

% Our final corpus statistics are reported in Table \ref{table:datasetStats}. 

%Since in this work, we created a two-fold 70/30\% train/test splits experimental setup, therefore, the numbers reported in train and test columns are averages. This two-fold experimental setup is explained in detail later in \autoref{eval}. Furthermore, in the first main column, i.e. the ``Our corpus'' column, when compared with the corpus from existing work by \citet{hou-etal-2019-identification}, i.e. the ``Prior work'' column, our corpus shows itself to be significantly larger thus showing a more large-scale evaluation setting.

We phrased the following question to formulate our task objective w.r.t. the (T, D, M, S) extraction target: \textit{What are the values for the following properties to construct a Leaderboard for the model introduced in this article: task, dataset, metric, and score?} In essence, it encapsulates an IE task.

Instruction tuning~\cite{unifiedqa,unifiedskg,supernaturalinstructions,p3,honovich2022unnatural,flan-collection} boosts LLMs by providing specific finetuning instructions, improving adaptability and performance on new tasks~\cite{instructgpt,flan-t5}. This method offers a more efficient approach than traditional unlabeled data methods~\cite{t5,liu2019multi,muppet,ext5}, allowing for versatile task prompting with single instructions.

In this vein, the ``Flan 2022 Collection''~\cite{flan-collection} was a large-scale open-sourced collection of 62 prior publicly released datasets in the NLP community clustered as 12 task types, such as reading comprehension (RC), sentiment, natural language inference (NLI), struct to text, etc. It is the most comprehensive resource facilitating open-sourced LLM development as generic multi-task models. Importantly, and of relevance to this work, FLAN was not just a super-amalgamation of datasets encapsulating different learning objectives, but also included at least \href{https://github.com/google-research/FLAN/blob/main/flan/templates.py}{10 human-curated natural instructions} per dataset that described the task for that dataset. As such, we select a set of instructions to guide the LLM for our complex IE task from the FLAN collection. Specifically, we identified the applicable instructions to our task were those designed for the SQuAD\_v2~\cite{squad-v1,squad-v2} and DROP~\cite{drop} datasets. Specifically, 8 SQuAD and 7 DROP instructions were found suitable. The general characteristic of the selected instructions is that they encode a context and the \textsc{SOTA} task objective, and instruct the model to fulfill the objective. The context, in our case, is a selection from specific sections from the full-text of an article where (T, D, M, S) information is most likely shared. This is discussed next.

As introduced in our prior work \cite{kabongo2024context}, we compared the performances of the four state-of-the-art LLMs on the following three contexts:

\subsection{DocTAET}

\begin{figure}[!th]
\centering
\includegraphics[width=16cm]{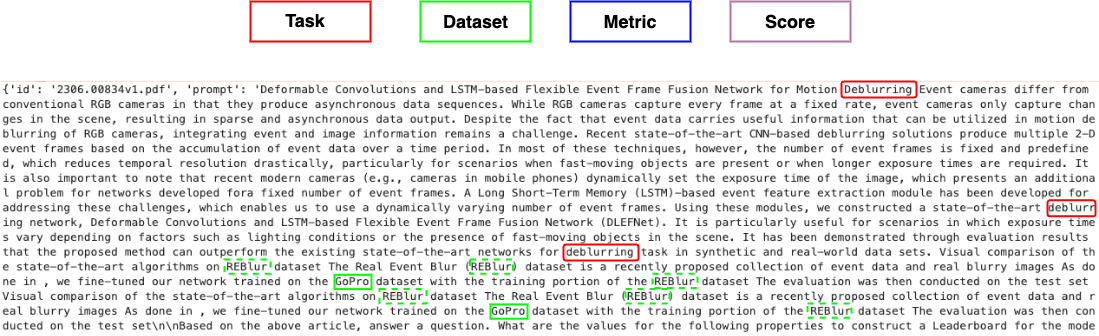}
\caption{DocTAET representation of the paper title \textit{"Deformable Convolutions and LSTM-based Flexible Event Frame Fusion Network for Motion Deblurring"} With Dashed lines representing Task, Dataset, Metrics, and Score present in the paper but not captured by paper with codes}
\label{fig:docteat-example}
\end{figure}

Delineated in prior work~\cite{hou-etal-2019-identification}, the context to the LLM comprises text selected from the (T)-title, (A)-abstract, (E)-experimental setup, and (T)-tabular information parts of the full-text. It yields an average context length of 493 words, ranging from a minimum of 26 words to a maximum of 7,361 words. These specific selections targeted the areas of the paper where the (T, D, M, S) were most likely to be found. An example of this context selection is illustrated in ~\autoref{fig:docteat-example}.

% As shown in Figure \ref{fig:docteat-example}

\subsection{DocREC}
\begin{figure}[!htb]
\centering
\includegraphics[width=16cm]{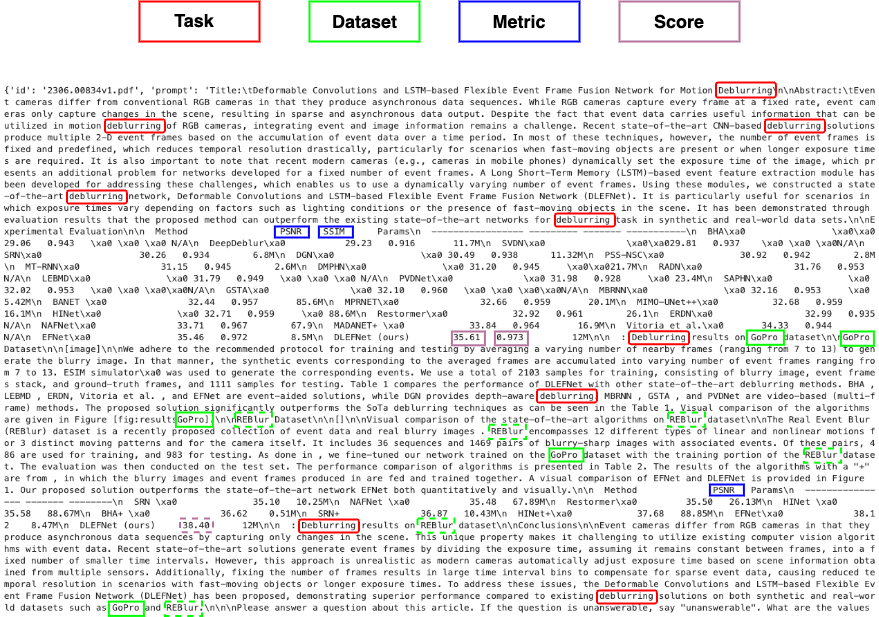}
\caption{DocREC representation of the paper title \textit{"Deformable Convolutions and LSTM-based Flexible Event Frame Fusion Network for Motion Deblurring"}. With Dashed lines representing Task, Dataset, Metrics, and Score present in the paper but not captured by paper with codes}
\label{fig:docrec-example}
\end{figure}

Introduced for the first time in this work, the DocREC context comprises text selected from the sections named (R)-results, (E)-experiments, and (C)-conclusions with allowances for variations in the three names. Complementary but still unique to DocTAET, the DocREC context representation aims to distill the essence of the research findings and conclusions into a succinct format. This context, ended by being much longer than DocTAET, yielded an average length of 1,586 words, with a minimum length of 27 words and a maximum length of 127,689 words. An example of this context selection is illustrated in ~\autoref{fig:docrec-example}.

\subsection{DocFULL}
As a last representation and to test the hypothesis that longer contexts which are not selectively tailored to suit the desired task objective tends to distract and thereby hinder the LLM performance, we used the full paper text as context. This approach entailed compiling the LaTeX source code of the document and translating its entirety into a plain text file. DocFULL ended producing the longest contexts compared to DocTAET and DocREC, in an average length of 5,948 words.

\subsection{Models}

In the period following the introduction of Flan-T5~\cite{flan-t5}, the community has witnessed the emergence of multiple advanced LLMs that have outperformed it. In this paper, we adopt the Flan-T5 fine-tuning paradigm and implement it in a comparative experimental setup on two of the latest state-of-the-art LLMs featured on the  \href{https://huggingface.co/spaces/HuggingFaceH4/open_llm_leaderboard}{public LLM leaderboard}.

In this work, we selected two open-sourced models as well as two closed-source models.

\noindent{\textbf{Mistral-7B.}} As the first open-sourced model, we selected Mistral-7B~\cite{mistral_v01}. This model as the name suggested is a 7-billion-parameter language model optimized for performance and efficiency. It introduces the Grouped-Query Attention (GQA) for rapid inference and reduced memory requirements, and Sliding Window Attention (SWA) for handling long sequences with lower computational costs. The model surpasses existing models in benchmarks, including reasoning, mathematics, and code generation tasks. It also features fine-tuning capabilities for instruction following, achieving superior performance in human and automated benchmarks. \href{https://huggingface.co/mistralai}{Mistral 7B} is designed for real-time applications, supports high throughput, and its architecture enables effective sequence generation with optimized memory usage. The model is released under the Apache 2.0 license, with its \href{https://github.com/mistralai/mistral-src}{source code} on Github, facilitating broad accessibility and application in various tasks.

\noindent{\textbf{Llama-2 7B.}} As the second open-sourced model, we selected the LLama-2 model~\cite{touvron2023llama}. The Llama-2 model is a collection of LLMs that range from 7 billion to 70 billion parameters, designed for both general and dialogue-specific applications. From the three available \href{https://huggingface.co/meta-llama}{Llama-2 model checkpoints}, i.e. 7B, 13B, and 70B, for comparability with our first select model, i.e. Mistral-7B, we choose the Llama-2 7B model. The Llama-2 models are fine-tuned for enhanced dialogue use cases and exhibit improved performance over existing open-source models in terms of helpfulness and safety, based on benchmarks and human evaluations. The Llama-2 family includes models optimized for different scales and introduces safety and fine-tuning methodologies to advance the responsible development of LLMs.

\noindent{\textbf{GPT-4-Turbo.}} As the first proprietary model, we leveraged GPT-4-Turbo, developed by OpenAI. GPT-4-Turbo is designed to offer enhanced performance and efficiency, providing faster response times and reduced computational costs. This model excels in various natural language processing tasks, including text generation, translation, summarization, and question answering. Optimized for high-throughput applications, GPT-4-Turbo maintains a high level of accuracy and consistency, making it suitable for real-time AI applications. Additionally, the model incorporates improvements in handling extensive contextual inputs, which enhances its capability in generating coherent and contextually relevant outputs.

\noindent{\textbf{GPT-4.o.}} For the proprietary model, we leveraged GPT-4.o, developed by OpenAI. GPT-4.o (“o” for “omni”) is designed to enable more natural human-computer interactions, accepting a wide range of input types including text, audio, image, and video, and generating outputs in text, audio, and image formats. The model is optimized for multimodal tasks, offering a significant advancement in vision and audio understanding compared to previous models.

We employed the QLORA (Quantum-enhanced Learning Optimization for Robust AI) \cite{dettmers2024qlora} framework for fine-tuning our open-sourced models, leveraging its advanced optimization capabilities to enhance model performance. QLORA has been recognized for its innovative approach to integrating quantum computing principles with machine learning, offering a novel pathway to overcoming traditional optimization challenges.

\section{Evaluations}
\label{eval}

\noindent{\textbf{Experimental setup.}} For training, we had one main experimental setting based on the 15 instructions. As elicited earlier in \autoref{sec:dataset}, each of the 15 instruction were instantiated with the 7,987 (T, D, M, S) data instances and the SOTA question resulting in a total of 119,805 instances to instruction finetune out LLMs. In this scenario, we hypothesized that this repetition in the data instances across the instructions would cause the resulting model to overfit the training dataset. Thus to control for this, we applied the following experimental setup. Each instruction was instantiated with a random selection of only half the 7,987 (T, D, M, S) data instances resulting in a finetuning dataset of a sizeable 41,340 instances that had leaderboards. The papers w/o leaderboards were also similarly halved. As shown in \autoref{table:DocREC_Stats}, the few-shot test included 1,401 paper instances, comprising 648 papers without leaderboards and 735 papers with leaderboards. The zero-shot test set consisted of 789 papers, with 548 papers without leaderboards and 241 papers with leaderboards. Model hyperparameter details are in \autoref{app:hyp}. In terms of compute, all experiments including inference were run on an NVIDIA 3090 GPU. Training took ~20 hours on the 50\% sampled dataset, while inference lasted 10 minutes for ~2k test instances.

% Section 4, Page 6+7:
% The corpora statistics seem a bit off. Some numbers for the elements (652, 994, 2353), do not appear in Table 1. The Few Shot dataset has 753+648=1401 elements, and the Zero Shot dataset has 241+548=789 elements, which totals to 2190 elements according to Github.
\begin{table*}[!htb]
  \centering
  \begin{adjustbox}{width=1\textwidth}
  
    \begin{tabular}{|p{2cm}|p{1.1cm}rrrr|rrrrr|l|}
      % \hline
      % & & \multicolumn{5}{|c|}{Highest Scores} & \multicolumn{5}{c|}{Lowest Scores} \\
      \toprule
      &        \multicolumn{5}{|c|}{Few-shot} & \multicolumn{5}{c}{Zero-shot} &     \\
      \toprule
    \bf Model & \bf Rouge1 & \bf Rouge2 & \bf RougeL & \bf RougeLsum & \bf \stackbox[c]{General\\-Accuracy} &\
       \bf  Rouge1 & \bf Rouge2 & \bf RougeL & \bf RougeLsum & \bf \stackbox[c]{General\\-Accuracy} & \bf Context \\
      \midrule
    \multirow{3}{*}{Llama-2 7B} & 49.68&10.18 &48.91 &49.02 &83.51 & 68.15& 4.81& 67.59& 67.78& 86.82& DocREC \\ \cline{2-12} 
      &49.70 &17.62 &48.81 &48.81 &83.62 & 62.75& 10.88& 62.07& 62.18& 86.22& DocTAET \\ \cline{2-12}
      &5.38 &0.79 &4.96 &5.13 &57.54 &7.55 &0.71 &7.24 &7.35 &37.80 &  DocFULL \\ \Xhline{2\arrayrulewidth}
    \multirow{3}{*}{Mistral 7B}  & 55.46& 14.11& 54.54& 54.64& 88.44& 72.98& 6.87& 72.42&72.35 & 92.40&  DocREC \\ \cline{2-12} 
      & 57.24& \textbf{19.67}& 56.28& \textbf{56.19}& \textbf{89.68}&73.54 &\textbf{12.23} & 73.01&72.95 &\textbf{95.97} &  DocTAET \\ \cline{2-12}
     & 6.73	&	0.77&	6.36&	6.49& 58.45&9.38 	&0.59	&9.11	&9.23	& 39.28& DocFULL \\ \Xhline{2\arrayrulewidth}
     \multirow{3}{*}{GPT-4-Turbo}  & 52.64&5.82 & 51.99&51.76 &60.89 & 72.80& 2.66& 72.35& 72.09& 77.06&   DocREC \\ \cline{2-12} 
        & 43.14 & 2.41&42.97 & 42.91& 47.33& 59.98& 0.48& 59.89&59.74  & 61.18&    DocTAET \\ \cline{2-12}
       & 48.50& 3.21& 48.06&47.96 & 52.87& 70.10& 1.8& 69.75&  69.73& 72.65&    DocFULL \\ \Xhline{2\arrayrulewidth}
     \multirow{3}{*}{GPT-4.o}  & \textbf{58.59} & 16.81& \textbf{56.37}& 55.45& 83.21& \textbf{74.94}& 9.02& \textbf{73.65}& \textbf{73.02}& 87.94&   DocREC \\ \cline{2-12} 
      &52.10 & 13.72& 50.77& 49.26& 80.63& 69.59& 8.81&	68.57& 67.43& 87.56& DocTAET \\ \cline{2-12}
       & 55.41 & 17.82& 53.01&51.79 & 79.56& 70.05& 10.89& 68.42& 67.51& 78.95&    DocFULL \\ \Xhline{2\arrayrulewidth}
    \end{tabular}
  \end{adjustbox}
  \caption{Evaluation results of Llama-2, Mistral, GPT-4-Turbo, and GPT-4.0 for the shared task, reported using the metrics proposed for the task. The output evaluations are conducted as a structured summary generation task (reported with ROUGE metrics) and as a binary classification task to distinguish between papers with and without leaderboards (reported as General Accuracy).}
  \label{tab:rouge-50percent}
\end{table*}

\noindent{\textbf{Metrics.}} We evaluated our models in two main settings. In the first setting, we applied standard summarization ROUGE metrics~\cite{rouge}. Furthermore, we also tested the models ability to identify papers with leaderboards and those without. This task was simple. For the papers with leaderboards, the model replied with a structured summary and for those it identified as without it replied as ``unanswerable.'' For these evaluations we applied simple accuracy measure. In the second setting, we evaluated the model JSON output in a fine-grained manner w.r.t. each of the inidividual (T, D, M, S) elements and overall for which we reported the results in terms of the standard F1 score and Precision score.

\begin{table*}[!htb]
\centering
\begin{adjustbox}{width=1\textwidth}
	\begin{tabular}{|l|l|ccccc|ccccc|l|} \hline
		&         & \multicolumn{5}{|c|}{Few-shot} & \multicolumn{5}{c}{Zero-shot} &     \\ \toprule
	{\bf Model}	& \bf Mode & \stackbox[c]{\bf Task}  & \bf Dataset & \bf Metric & \stackbox[c]{\bf Score} & \bf Overall & \stackbox[c]{\bf Task} & \bf Dataset & \bf Metric & \stackbox[c]{\bf Score} & \bf Overall & \bf Context \\ \Xhline{2\arrayrulewidth}
        \multirow{6}{*}{Llama-2 7B} & \textcolor{orange}{Exact} &20.93 & 13.06&13.96 &3.04 &12.75 & 13.97& 6.83& 11.72& 2.61& 8.78& \multirow{2}{*}{DocREC} \\ \cline{2-12}
		& \textcolor{blue}{Partial} & 31.37& 22.50&21.99 &3.46 &19.83 & 24.05& 16.6&18.28 &3.10 &15.51 & \\ \cline{2-13}
		\multirow{2}{*}{} & \textcolor{orange}{Exact}  & 29.53 & 16.68 & 20.02 & 1.14 & 16.84 & 21.75 & 11.26 & 16.99 & 0.77 & 12.69 &  \multirow{2}{*}{DocTAET} \\ \cline{2-12}
		& \textcolor{blue}{Partial} & 43.37 & 30.36 & 30.51 & 1.38 & 26.40 & 38.48 & 23.10 & 27.09 & 0.96 & 22.41 & \\ \cline{2-13}
		\multirow{2}{*}{} & \textcolor{orange}{Exact} & 1.59 & 1.36& 0.94& 0.23& 1.03& 2.06 &1.30 &1.52 &0.33 &1.30 & \multirow{2}{*}{DocFULL} \\ \cline{2-12}
		& \textcolor{blue}{Partial}  &2.29 &1.82 &1.68 &0.37 &1.54 & 3.36 &2.49 &2.49 &0.54 &2.22 & \\ \Xhline{2\arrayrulewidth}
        \multirow{6}{*}{Mistral 7B} & \textcolor{orange}{Exact}  & 26.77 & 15.68& 18.70& 6.36& 16.88&17.99 & 11.80& 15.55& 5.04& 12.60&  \multirow{2}{*}{DocREC} \\ \cline{2-12}
		& \textcolor{blue}{Partial}  &39.75 &27.28 &28.49 &7.08 & 25.65& 29.88&21.05 &23.16 & 5.75& 19.96& \\ \cline{2-13}
		\multirow{2}{*}{} & \textcolor{orange}{Exact}  & \textbf{33.38} & \textbf{18.51}& \textbf{24.23}& 1.87& \textbf{19.50}& \textbf{26.99}& 14.32& \textbf{22.04}& 1.20& \textbf{16.14} & \multirow{2}{*}{DocTAET} \\ \cline{2-12}
		& \textcolor{blue}{Partial}  & \textbf{46.35}& \textbf{32.75}& \textbf{34.16}& 2.25& 28.88& \textbf{44.90}& 27.29& \textbf{32.23}& 1.41& 26.46& \\ \cline{2-13}
		\multirow{2}{*}{} & \textcolor{orange}{Exact} & 0.81& 0.57& 0.57& 0.56& 0.63&0.22 &0.33 &0.33 &0.76 & 0.42&  \multirow{2}{*}{DocFULL} \\ \cline{2-12}
		& \textcolor{blue}{Partial} & 1.19& 0.85& 0.81& 0.84& 0.92& 0.56& 0.67& 0.78& 0.87& 0.72&  \\ \Xhline{2\arrayrulewidth}
        \multirow{6}{*}{GPT-4-Turbo} & \textcolor{orange}{Exact}   & 7.61& 6.19& 4.92& 4.25& 5.74& 4.26& 5.35& 3.86& 3.28& 4.18& \multirow{2}{*}{DocREC} \\ \cline{2-12}
		& \textcolor{blue}{Partial}   & 16.48& 13.96& 11.03& 7.03& 12.13& 13.76& 11.09& 10.19& 5.46& 10.13& \\ \cline{2-13}
		\multirow{2}{*}{} & \textcolor{orange}{Exact} & 2.99& 2.69& 0.95& 0.75& 1.84& 1.13& 0.79& 0.34& 0.11& 0.59& \multirow{2}{*}{DocTAET} \\ \cline{2-12}
		& \textcolor{blue}{Partial} & 6.22& 5.42& 3.03& 1.63& 4.08& 2.72& 1.59& 1.59& 0.11& 1.5& \\ \cline{2-13}
		\multirow{2}{*}{} & \textcolor{orange}{Exact}  & 3.38& 3.16& 1.98& 2.48& 2.75& 2.45& 2.98& 1.81& 2.77& 2.5&  \multirow{2}{*}{DocFULL} \\ \cline{2-12}
		& \textcolor{blue}{Partial}  & 7.03& 6.41& 4.96& 4.15& 5.64& 6.49& 5.85& 4.47& 3.56& 5.09&  \\ \Xhline{2\arrayrulewidth}
        \multirow{6}{*}{GPT-4.o} & \textcolor{orange}{Exact}  & 16.14& 16.11& 15.50& 10.76& 14.63& 16.04& \textbf{15.05}& 17.43& 10.38& 14.72&  \multirow{2}{*}{DocREC} \\ \cline{2-12}
		& \textcolor{blue}{Partial}  & 38.40& 32.63& 29.35& 15.20& \textbf{28.90}& 37.23& \textbf{31.16}& 29.97& \textbf{14.96}& \textbf{28.33}& \\ \cline{2-13}
		\multirow{2}{*}{} & \textcolor{orange}{Exact}  &14.10 &12.76 &9.91 &2.11 & 9.72& 13.78& 10.25& 11.01& 2.36& 9.35& \multirow{2}{*}{DocTAET} \\ \cline{2-12}
		& \textcolor{blue}{Partial}  & 31.84& 26.65& 20.83& 4.22& 20.92& 29.33& 23.87& 19.50& 3.71& 19.12& \\ \cline{2-13}
		\multirow{2}{*}{} & \textcolor{orange}{Exact}   &16.72& 14.53& 14.67& \textbf{11.25}& 14.29& 13.08& 14.94&16.09&\textbf{11.17}& 13.82&  \multirow{2}{*}{DocFULL} \\ \cline{2-12}
		& \textcolor{blue}{Partial} & 36.56& 31.0&27.61&\textbf{16.50}&27.93&35.59&28.28&27.38&14.80&26.52& \\ \Xhline{2\arrayrulewidth}
	\end{tabular}
\end{adjustbox}
\caption{Evaluation results of Llama-2, Mistral, GPT-4-Turbo, and GPT-4.0 for the shared task, reported using the metrics proposed for the task. The evaluation considers the individual (Task, Dataset, Metric, Score) elements and Overall in the model JSON generated output, reported in terms of \textbf{F1 score}.}
\label{tab:f1-50percent}
\end{table*}

\begin{table*}[!htb]
\centering
\begin{adjustbox}{width=1\textwidth}
	\begin{tabular}{|l|l|ccccc|ccccc|l|} \hline
		&         & \multicolumn{5}{|c|}{Few-shot} & \multicolumn{5}{c}{Zero-shot} &     \\ \toprule
	{\bf Model}	& \bf Mode & \stackbox[c]{\bf Task}  & \bf Dataset & \bf Metric & \stackbox[c]{\bf Score} & \bf Overall & \stackbox[c]{\bf Task} & \bf Dataset & \bf Metric & \stackbox[c]{\bf Score} & \bf Overall & \bf Context \\ \Xhline{2\arrayrulewidth}
        \multirow{6}{*}{Llama-2 7B} & \textcolor{orange}{Exact}  & 34.10& 21.27& 22.74& 4.99& 20.78& 31.89& 15.77& 26.77& 6.06& 20.12&  \multirow{2}{*}{DocREC} \\ \cline{2-12}
		& \textcolor{blue}{Partial} & 51.13& 36.66& 35.82& 5.59& 32.32& 54.92& 38.32& 41.73& 7.27& 35.56&  \\ \cline{2-13}
		\multirow{2}{*}{} & \textcolor{orange}{Exact}  & 30.61& 17.29& 20.78& 1.18& 17.46& 24.56& 12.72& 19.19& 0.87& 14.34&  \multirow{2}{*}{DocTAET} \\ \cline{2-12}
		& \textcolor{blue}{Partial}  & 44.96& 31.48& 31.66& 1.43& 27.38& 43.46& 26.09& 30.60& 1.09& 25.31&  \\ \cline{2-13}
		\multirow{2}{*}{} & \textcolor{orange}{Exact}  & 34.69& 29.59& 20.41& 5.10& 22.45& 32.20& 20.34& 23.73& 5.08& 20.34&  \multirow{2}{*}{DocFULL} \\ \cline{2-12}
		& \textcolor{blue}{Partial}  & 50.00& 39.80& 36.73& 8.16& 33.67& 52.54& 38.98& 38.98& 8.47& 34.75&  \\ \Xhline{2\arrayrulewidth}
        \multirow{6}{*}{Mistral 7B} & \textcolor{orange}{Exact}   & 37.65& 22.15& 26.38& 8.94& 23.78& 35.68& 23.40& 31.02& 9.98& 25.02&  \multirow{2}{*}{DocREC} \\ \cline{2-12}
		& \textcolor{blue}{Partial}  &55.90 & 38.52& 40.18& 9.95& 36.14& 59.25& 41.73& 46.20& 11.46& 39.66& \\ \cline{2-13}
		\multirow{2}{*}{} & \textcolor{orange}{Exact}   & \textbf{39.48}& 21.89& 28.66& 2.21& 23.06& \textbf{38.46}& 20.41& 31.41& 1.71& 23.00&  \multirow{2}{*}{DocTAET} \\ \cline{2-12}
		& \textcolor{blue}{Partial} & 54.82& 38.73& 40.41& 2.65& 34.15& 64.00& 38.89& 45.94& 2.03& 37.71&  \\ \cline{2-13}
		\multirow{2}{*}{} & \textcolor{orange}{Exact}  & 32.43& \textbf{32.43}& \textbf{32.43}& 9.6& \textbf{30.76}&25.00 &\textbf{37.50} &\textbf{37.50} &14.00 & \textbf{28.50}&   \multirow{2}{*}{DocFULL} \\ \cline{2-12}
		& \textcolor{blue}{Partial} & \textbf{71.43}& 48.65& \textbf{45.95}& 14.52& \textbf{45.13}&62.50 &\textbf{75.00} &\textbf{87.50} &21.62 & \textbf{61.66}& \\ \Xhline{2\arrayrulewidth}
        \multirow{6}{*}{GPT-4-Turbo} & \textcolor{orange}{Exact}   & 30.46& 24.80& 19.67& 17.30& 23.06& 20.57& 25.96& 18.66& 15.94& 20.28&  \multirow{2}{*}{DocREC} \\ \cline{2-12}
		& \textcolor{blue}{Partial} & 65.96& 55.95& 44.07& 23.89& 47.47& 66.51& 53.85& 49.28& 21.67& 47.83& \\ \cline{2-13}
		\multirow{2}{*}{} & \textcolor{orange}{Exact} & 32.79& 29.51& 10.27& 8.20& 20.19& 29.41& 20.59& 8.82& 2.94& 15.44& \multirow{2}{*}{DocTAET} \\ \cline{2-12}
		& \textcolor{blue}{Partial} & 68.31& \textbf{59.56}& 32.97& 15.64& 44.12& \textbf{70.59}& 41.18& 41.18& 4.55& 39.37& \\ \cline{2-13}
		\multirow{2}{*}{} & \textcolor{orange}{Exact} & 28.52& 27.56& 17.05& \textbf{21.65}& 23.70& 24.73& 30.11& 18.28& \textbf{27.96}& 25.73&  \multirow{2}{*}{DocFULL} \\ \cline{2-12}
		& \textcolor{blue}{Partial} & 59.32& 55.91& 42.64& \textbf{30.69}& 47.14& 65.59& 59.14& 45.16& \textbf{30.91}& 50.20& \\ \Xhline{2\arrayrulewidth}
        \multirow{6}{*}{GPT-4.o} & \textcolor{orange}{Exact}  & 17.99& 17.96& 17.28& 12.00& 16.31& 19.5& 18.72& 21.67& 12.89& 18.31&  \multirow{2}{*}{DocREC} \\ \cline{2-12}
		& \textcolor{blue}{Partial}  & 42.82& 36.38& 32.72& 16.09& 32.00&46.31 &38.5 &37.27 &18.18 &35.13 & \\ \cline{2-13}
		\multirow{2}{*}{} & \textcolor{orange}{Exact}  & 19.70& 17.83& 13.85& 2.95&13.58 & 25.31& 18.83&20.22 &4.37 &17.18 & \multirow{2}{*}{DocTAET} \\ \cline{2-12}
		& \textcolor{blue}{Partial}  & 44.50& 37.25& 29.11& 7.78& 29.66&53.86 &43.83 &35.80 &9.07 &35.64 & \\ \cline{2-13}
		\multirow{2}{*}{} & \textcolor{orange}{Exact} & 19.18& 16.67& 16.84&12.90& 16.40& 15.30& 17.48&18.83&12.05&16.17&  \multirow{2}{*}{DocFULL} \\ \cline{2-12}
		& \textcolor{blue}{Partial} & 41.95& 35.56&31.70&18.15&31.84&41.64&33.08&32.03&16.66&30.85& \\ \Xhline{2\arrayrulewidth}
	\end{tabular}
\end{adjustbox}
\caption{Evaluation results of Llama-2, Mistral, GPT-4-Turbo, and GPT-4.o w.r.t. the individual (Task, Dataset, Metric, Score) elements and Overall in the model JSON generated output in terms of \textbf{Precision score}.}
\label{tab:precision-50percent}
\end{table*}

\begin{table*}[!htb]
\centering
\begin{adjustbox}{width=1\textwidth}
	\begin{tabular}{|l|l|ccccc|ccccc|l|} \hline
		&         & \multicolumn{5}{|c|}{Few-shot} & \multicolumn{5}{c}{Zero-shot} &     \\ \toprule
	{\bf Model}	& \bf Mode & \stackbox[c]{\bf Task}  & \bf Dataset & \bf Metric & \stackbox[c]{\bf Score} & \bf Overall & \stackbox[c]{\bf Task} & \bf Dataset & \bf Metric & \stackbox[c]{\bf Score} & \bf Overall & \bf Context \\ \Xhline{2\arrayrulewidth}
        \multirow{6}{*}{Llama-2 7B} & \textcolor{orange}{Exact}  & 15.09& 9.42& 10.07& 2.19& 9.19&8.94 & 4.36& 7.51& 1.66& 5.62&  \multirow{2}{*}{DocREC} \\ \cline{2-12}
		& \textcolor{blue}{Partial}  & 2.63& 16.23& 15.86& 2.49& 14.30& 15.40& 10.60& 11.70& 1.97& 9.92&  \\ \cline{2-13}
		\multirow{2}{*}{} & \textcolor{orange}{Exact}  & 28.52& \textbf{16.11}& 19.32& 1.10& 16.26& 19.52& 10.10& 15.24& 0.69& 11.39&  \multirow{2}{*}{DocTAET} \\ \cline{2-12}
		& \textcolor{blue}{Partial}  & 41.89& 29.33& 29.43& 1.33& 25.50& 34.53& 20.73& 24.31& 0.85& 20.10&  \\ \cline{2-13}
		\multirow{2}{*}{} & \textcolor{orange}{Exact}  & 0.81& 0.69& 0.48& 0.12& 0.53& 1.06& 0.67& 0.78& 0.17& 0.67&  \multirow{2}{*}{DocFULL} \\ \cline{2-12}
		& \textcolor{blue}{Partial}  & 1.17& 0.93& 0.86& 0.19& 0.79& 1.74& 1.29& 1.29& 0.28& 1.15&  \\ \Xhline{2\arrayrulewidth}
        \multirow{6}{*}{Mistral 7B} & \textcolor{orange}{Exact} & 20.77& 12.14& 14.49& 4.93& 13.08& 12.03& 7.89& 10.38& 3.37& 8.42&   \multirow{2}{*}{DocREC} \\ \cline{2-12}
		& \textcolor{blue}{Partial}  & 30.84& 21.12& 22.07& 5.49& 19.88& 19.98& 14.07& 15.45& 3.84& 13.33&  \\ \cline{2-13}
		\multirow{2}{*}{} & \textcolor{orange}{Exact}  & \textbf{28.91}& 16.03& \textbf{20.99}& 1.62& \textbf{16.89}& \textbf{20.79}& 11.03& \textbf{16.97}& 0.93& \textbf{12.43}&  \multirow{2}{*}{DocTAET} \\ \cline{2-12}
		& \textcolor{blue}{Partial}  & \textbf{40.15}& 28.36& \textbf{29.59}& 1.95& 25.01& \textbf{34.58}& 21.02& 24.83& 1.08& 20.38&  \\ \cline{2-13}
		\multirow{2}{*}{} & \textcolor{orange}{Exact}  &0.41 &0.29 &0.29 &0.29 &0.32 & 0.11& 0.17& 0.17& 0.39&0.21 &    \multirow{2}{*}{DocFULL} \\ \cline{2-12}
		& \textcolor{blue}{Partial}  & 0.6& 0.43& 0.41& 0.43& 0.47& 0.28& 0.34& 0.39& 0.44& 0.36&  \\ \Xhline{2\arrayrulewidth}
        \multirow{6}{*}{GPT-4-Turbo} & \textcolor{orange}{Exact}  & 4.35& 3.53& 2.81& 2.42& 3.28& 2.37& 2.98& 2.15& 1.83& 2.33&  \multirow{2}{*}{DocREC} \\ \cline{2-12}
		& \textcolor{blue}{Partial}  & 9.42& 7.98& 6.30& 4.12& 6.95& 7.67& 6.18& 5.68& 3.12& 5.66&  \\ \cline{2-13}
		\multirow{2}{*}{} & \textcolor{orange}{Exact}  &1.56 &1.41 &0.50 &0.39 &0.96 & 0.58& 0.40& 0.17& 0.06& 0.30&  \multirow{2}{*}{DocTAET} \\ \cline{2-12}
		& \textcolor{blue}{Partial}  &3.26 &2.84 &1.59 &0.86 &2.14 & 1.39& 0.81& 0.81& 0.06& 0.76&  \\ \cline{2-13}
		\multirow{2}{*}{} & \textcolor{orange}{Exact}  & 1.8& 1.68& 1.05& 1.32& 1.46& 1.29& 1.57& 0.95& 1.46& 1.32&  \multirow{2}{*}{DocFULL} \\ \cline{2-12}
		& \textcolor{blue}{Partial} & 3.73& 3.40& 2.63& 2.23& 3.00& 3.42& 3.08& 2.35& 1.89& 2.68&  \\ \Xhline{2\arrayrulewidth}
        \multirow{6}{*}{GPT-4.o} & \textcolor{orange}{Exact} & 14.63& 14.60& 14.05& 9.75& 13.26& 13.41& \textbf{12.58}& 14.57& 8.68& 12.31&  \multirow{2}{*}{DocREC} \\ \cline{2-12}
		& \textcolor{blue}{Partial}  & 34.81& \textbf{29.58}& 26.60& 14.41& \textbf{26.35}& 31.13& \textbf{26.05}& \textbf{25.06}& 12.71& \textbf{23.74}& \\ \cline{2-13}
		\multirow{2}{*}{} & \textcolor{orange}{Exact}  & 10.97& 9.93& 7.72& 1.64& 7.57& 9.47& 7.04& 7.56& 1.62& 6.42&  \multirow{2}{*}{DocTAET} \\ \cline{2-12}
		& \textcolor{blue}{Partial}   & 24.79& 20.75& 16.21& 2.89& 16.16& 20.15& 16.40& 13.39& 2.33& 13.07&  \\ \cline{2-13}
		\multirow{2}{*}{} & \textcolor{orange}{Exact}  & 14.82& 12.88& 13.00& \textbf{9.97}& 12.66& 11.42& 13.05& 14.05& \textbf{9.76}& 12.07&   \multirow{2}{*}{DocFULL} \\ \cline{2-12}
		& \textcolor{blue}{Partial} & 32.41& 27.48& 24.46& \textbf{15.13}& 24.87& 31.08& 24.69& 23.91& \textbf{13.32}& 23.25& \\ \Xhline{2\arrayrulewidth}
	\end{tabular}
\end{adjustbox}
\caption{Evaluation results of Llama-2, Mistral, GPT-4-Turbo, and GPT-4.o w.r.t. the individual (Task, Dataset, Metric, Score) elements and Overall in the model JSON generated output in terms of \textbf{Recall score}.}
\label{tab:recall-50percent}
\end{table*}

\subsection{Results and Discussion}

This section analyzes the comparative performance of different contexts provided to LLMs, examining their impact on model precision and reliability across various tasks and settings. Additionally, we utilized a matching algorithm, fuzz.ratio, with a threshold of 50\% for partial metrics to account for variations in how tasks, datasets, and metrics are reported in research papers compared to the pwc code dumps.\\

\noindent{\textbf{Results for RQ1: Performance in Generating Structured Summaries and Classification}}

\begin{enumerate}
    \item Few-shot Performance (\autoref{tab:rouge-50percent}):
    \begin{itemize}
        \item \textbf{Llama-2 7B:} Demonstrates consistent performance across different contexts (DocREC, DocTAET) with high General Accuracy in DocREC (83.51\%) and DocTAET (83.62\%). It shows substantial differences in ROUGE scores, especially in the ROUGE-2 and ROUGE-Lsum metrics, indicating its ability to capture more nuanced information.
        \item \textbf{Mistral 7B:} Exhibits the highest General Accuracy in DocTAET (89.68\%) and competitive ROUGE scores across all contexts. This model shows significant improvement over Llama-2 7B and GPT-4-Turbo in most metrics, showcasing the strengths of domain-specific fine-tuned open-source models.
        \item 	\textbf{GPT-4-Turbo:} Shows moderate performance with lower General Accuracy in DocTAET (47.33\%). However, it demonstrates notable improvements in ROUGE-L and ROUGE-Lsum, indicating its potential in capturing detailed summaries.
        \item  \textbf{GPT-4.o:} Shows strong performance in ROUGE metrics, especially in DocREC and DocTAET contexts, with high General Accuracy in DocREC (83.21\%) and DocTAET (80.63\%). However, the DocFULL context results of 79.56\% which indicate the robustness of the DocREC context.
    \end{itemize}
    \item Zero-shot Performance (\autoref{tab:rouge-50percent}):
    \begin{itemize}
        \item \textbf{Llama-2 7B:} Performs well with notable General Accuracy in DocREC (86.82\%) and DocTAET (86.22\%). However, there is a significant drop in performance in the DocFULL context (37.8\% General Accuracy), likely due to the long and complex nature of the full document context which introduces distractions and challenges the model's ability to maintain focus and coherence over extended text. This suggests that while the models perform well in more concise contexts, the added complexity and length in DocFULL can detract from their overall accuracy and effectiveness. 
        \item \textbf{Mistral 7B:} Outperforms other models in General Accuracy, especially in DocTAET (95.97\%) and DocREC (92.40\%). This indicates its robustness in zero-shot settings, making it a reliable choice for generating structured summaries without prior examples.
        \item \textbf{GPT-4-Turbo:} Shows lower General Accuracy in zero-shot settings, particularly in DocREC (77.06\%) and DocTAET (61.18\%), highlighting the need for further optimization.
        \item  \textbf{GPT-4.o:} Continues to show competitive performance in DocREC (87.94\%) and DocTAET (87.56\%) contexts, reinforcing its capability in zero-shot scenarios.
    \end{itemize}
\end{enumerate}

\noindent{\textbf{Results for RQ2: Precision and Performance Trade-offs}

\begin{enumerate}
    \item Few-shot F1 Score analysis (\autoref{tab:f1-50percent}):
    \begin{itemize}
        \item \textbf{Llama-2 7B:} Shows a balanced performance between exact and partial metrics. For example, it achieves an Overall score of 26.40 in the partial mode under DocTAET context, indicating its ability to handle partial matches effectively.
        \item \textbf{Mistral 7B:} Excels in both exact and partial metrics, with the highest Overall score of 28.88 in the partial mode under DocTAET context. This model provides the best trade-off between precision and other performance metrics.
        \item \textbf{GPT-4-Turbo:} Shows moderate performance with an Overall score of 12.13 in the partial mode under DocREC context. It requires further optimization to compete with other models.
        \item  \textbf{GPT-4.o:} Demonstrates strong partial mode performance, with an Overall score of 28.90 under DocREC context.
    \end{itemize}
    \item Zero-shot F1 Score analysis (\autoref{tab:f1-50percent}):
    \begin{itemize}
        \item \textbf{Llama-2 7B:} Shows variability in performance with a notable Overall score of 22.41 in the partial mode under DocTAET context.
        \item \textbf{Mistral 7B:} Continues to lead with the highest Overall score of 16.14 in the exact mode under DocTAET context, reinforcing its capability in zero-shot scenarios.
        \item \textbf{GPT-4-Turbo:} Provides moderate performance in partial mode with an Overall score of 10.13 under DocFULL context, highlighting areas for improvement.
        \item  \textbf{GPT-4.o:} Provides solid performance in partial mode with an Overall score of 23.87 under DocTAET context.
    \end{itemize}
    
    \item Precision Scores (\autoref{tab:precision-50percent}):
    
    \begin{itemize}
        \item \textbf{Llama-2 7B:} Achieves moderate precision scores, particularly in the partial mode with an Overall score of 35.56 in zero-shot settings under DocREC context.
        \item \textbf{Mistral 7B:} Dominates precision metrics with the highest Overall scores in both exact and partial modes across different contexts, demonstrating its precision and reliability.
        \item \textbf{GPT-4-Turbo:} Shows competitive precision scores in zero-shot settings, particularly in the partial mode with an Overall score of 50.20 under DocFULL context.
        \item  \textbf{GPT-4.o:} Achieves moderate precision scores in few-shot settings, particularly in the exact mode, with an Overall score of 16.40 under DocFULL context.
    \end{itemize}
    
    \item Recall Scores (\autoref{tab:recall-50percent}):
    
    \begin{itemize}
        \item \textbf{Llama-2 7B:} Demonstrates notable recall scores in few-shot settings, particularly in the partial mode with an Overall score of 25.50 under DocTAET context. However, its performance varies significantly across different contexts and modes.
        \item \textbf{Mistral 7B:} Leads in recall scores, achieving the highest Overall scores in both exact and partial modes across various contexts. Notably, it achieves an Overall score of 16.89 in the exact mode in few-shot setting and 12.43 in few-shot under DocTAET context.
        \item \textbf{GPT-4-Turbo:} Shows moderate recall performance, with room for improvement, particularly in the zero-shot settings. Its best recall score is in the exact mode under DocREC context with an Overall score of 3.28.
        \item  \textbf{GPT-4.o:} Excels in recall scores, especially in few-shot settings, achieving an Overall score of 26.35 in the partial mode under DocREC context. It maintains competitive recall scores across different contexts and modes.
    \end{itemize}
\end{enumerate}

\noindent{\textbf{Score Extraction Analysis}}\\

When focusing on the extraction of score entities, GPT models, particularly GPT-4.o, demonstrate a clear advantage. As shown in~\autoref{tab:f1-50percent}, GPT-4.o outperforms other models in extracting scores with higher accuracy and precision. For example, in few-shot settings, GPT-4.o achieves an Overall score of 11.25 in the exact mode and 16.50 in the partial mode under the DocFULL context, significantly higher than Llama-2 7B and Mistral 7B. Similarly, in zero-shot settings, GPT-4.o maintains a competitive edge with an Overall score of 14.96 in the partial mode under DocREC context, surpassing the performance of the other models.

This superior performance in score extraction can be attributed to the advanced capabilities of GPT-4.o and its optimized architecture for handling diverse and long contexts. This makes it particularly effective in accurately recognizing and extracting numerical and textual data, which is crucial for generating accurate leaderboard metrics and scores.\\

\noindent{\textbf{Discussion}}\\

Open-source models like Mistral 7B demonstrate competitive and sometimes superior performance compared to proprietary models like GPT-4.o and GPT-4-Turbo. This is evident in both structured summary generation and classification tasks, particularly in the DocTAET context, where Mistral 7B consistently outperforms others.

The trade-off between precision and performance is well-balanced in DocREC on Mistral 7B, making it a reliable choice for applications requiring high precision. This is crucial for scholarly communications where accuracy and reliability are paramount.
	
Few-shot settings show that all models, including open-source ones, perform robustly, but Mistral 7B often leads in various metrics. This highlights the potential of fine-tuning open-source models to achieve high performance that surpasses closed-source models.

\section{Conclusions}

Our participation in the shared task has demonstrated that fine-tuning open-source models like Mistral 7B and Llama-2 7B can yield competitive, and in some cases superior, results compared to proprietary models such as GPT-4.o and GPT-4-Turbo. Throughout our experiments, the DocTAET context typically delivered dependable and accurate performance, while the DocREC context excelled in scenarios where precision is paramount. The interaction among different context types and their consequent impact on model performance provided valuable insights into ongoing research and practical deployment of LLMs.

The implications of our findings within the context of the shared task are substantial. They indicate that with meticulous context design and implementation, fine-tuned open-source LLMs are well-suited for tracking and synthesizing scientific progress. This capability enables the provision of current and nuanced leaderboards for any given academic field. The potential of this technology to support and augment the efforts of academics and policymakers is significant, heralding novel opportunities for automated or semi-automated leaderboard construction.

While we have made significant progress in elucidating and enhancing context selection for LLMs, several avenues for future exploration have emerged. Investigating hybrid context selection methods, domain-specific adaptations, and the integration of structured data could lead to even more sophisticated leaderboard generation. As LLMs continue to advance, optimizing their potential remains a dynamic and impactful field of research.

In conclusion, our involvement in the shared task has not only highlighted the effectiveness of fine-tuned open-source models but also emphasized the importance of strategic context selection in maximizing model performance. These insights contribute to the broader understanding of LLM capabilities and pave the way for future advancements in the field.

% Our findings highlight that fine-tuning open-source models like Mistral 7B and Llama-2 7B can yield competitive, and in some cases superior, results compared to the proprietary model GPT-4.o and GPT-4-Turbo. The DocTAET context typically yields dependable and accurate performance, while the DocREC context excels in scenarios where precision is paramount. The interaction among different context types and their consequent impact on model performance offers extensive insights for ongoing research and the practical deployment of language models.

% The ramifications of our research are substantial, indicating that with meticulous context design and implementation, fine-tuned open-source LLMs are well-equipped for tracking and synthesizing scientific progress, thus providing a current and nuanced leaderboard for any given academic field. The potential of this technology to support and augment the endeavors of academics and policymakers is significant, heralding novel opportunities for automated, real-time meta-analysis.

% While this research makes significant progress in elucidating and enhancing
% context selection for LLMs, it also unveils several avenues for future exploration.
% Delving into hybrid context selection methods, domain-specific adaptations, and
% the assimilation of structured data could lead to even more sophisticated leader-
% board generation. As LLMs continue to advance, the pursuit of optimizing their
% potential remains a dynamic and impactful field of research.

\section*{Acknowledgements}

This work was jointly supported by the German BMBF project SCINEXT (01lS22070) and the Deutsche Forschungsgemeinschaft (DFG, German Research Foundation) – project number: NFDI4DataScience (460234259).

%%
%% Define the bibliography file to be used
\bibliography{main}

\appendix

% \section{Example Appendix}
% \label{sec:appendix}
\newpage
\section{Instructions: Qualitative Examples}
\label{app:instructions}

In this section, we elicit each of the instructions that were considered in this work as formulated in the FLAN 2022 Collection for the SQuAD\_v2 and DROP datasets.
\begin{table}[!h]
\centering
\resizebox{\textwidth}{!}{%
\begin{tabular}{@{}cll@{}}
\toprule
\textbf{ID} & \textbf{SQuAD\_v2 Instructions}                                      & \textbf{DROP Instructions}                                \\ \midrule
1  & Please answer a question about this article. If unanswerable, say "unanswerable". & Answer based on context.                                 \\
2  & \{Context\} \{Question\} If unanswerable, say "unanswerable".                     & Answer this question based on the article.               \\
3  & Try to answer this question if possible (otherwise reply "unanswerable").         & \{Context\} \{Question\}                                 \\
4  & Please answer a question about this article, or say "unanswerable" if not possible.& Answer this question: \{Question\}                       \\
5  & If possible to answer this question, do so (else, reply "unanswerable").          & Read this article and answer this question.              \\
6  & Answer this question, if possible (if impossible, reply "unanswerable").          & Based on the above article, answer a question.           \\
7  & What is the answer? (If it cannot be answered, return "unanswerable").            & Context: \{Context\} Question: \{Question\} Answer:      \\
8  & Now answer this question, if there is an answer (else, "unanswerable").           &                                                          \\ \bottomrule
\end{tabular}%
}
\caption{Comparative Instructions for the SQuAD\_v2 and DROP datasets.}
\label{table:datasets_instructions}
\end{table}

% \section{Our Experimental Hyperparamters}
% \label{app:hyp}
% We used two main experimental settings in this work. The first consists of a dataset of all 15 instruction templates, and the second one is a randomly selected half of every individual template instance in our dataset.

% Given that the average context length of our dataset was close to the 512 sequence length limit by T5 and the size of the available GPU, a batch size of 2 and gradient\_accumulation\_steps of 4 were used, which led to a final batch size of 8. All experiments were run on five epochs and we used AdafactorSchedule and Adafactor optimizer~\cite{shazeer2018adafactor} with scale\_parameter=True, relative\_step=True, warmup\_init=True, lr=None.

% The evaluations were all done on a dataset made of individual template instructions separately, as reported in table \ref{tab:rouge-50percent} to \ref{tab:f1-all}.

% The evaluations were done on each epoch on the dev set and we kept two best (the one maximizing the "Overall Partial F1" score) and last checkpoints in each model training process to then use for inference on test set.

\section{ROUGE Evaluation Metrics}
\label{app:rouge}

The ROUGE metrics~\cite{rouge} are commonly used for evaluating the quality of text summarization systems. ROUGE-1 measures the overlap of unigram (single word) units between the generated summary and the reference summary. ROUGE-2 extends this to measure the overlap of bigram (two consecutive word) units. ROUGE-L calculates the longest common subsequence between the generated and reference summaries, which takes into account the order of words. ROUGE-LSum is an extension of ROUGE-L that considers multiple reference summaries by treating them as a single summary. 
% These metrics provide a quantitative assessment of the similarity between the generated and reference summaries, helping researchers and developers evaluate and compare the effectiveness of different summarization approaches. They have become widely used benchmarks in the field of automatic summarization.

\section{Additional Data statistics and Hyperparameters}
% \label{app:data_stats}
\label{app:hyp}

\begin{table*}[h]
\begin{center}
\begin{threeparttable}
\begin{minipage}{\textwidth}
\begin{tabular*}{\textwidth}{@{\extracolsep{\fill}}l|ccc@{\extracolsep{\fill}}}
% \toprule%
\cmidrule{1-4}%
& \multicolumn{3}{@{}c@{}}{\textbf{Our Corpus}} \\\cmidrule{2-4}%
 & Train &Test-Few-shot & Test Zero-shot \\
\midrule
Papers w/ leaderboards & 7,744 & 961& 630  \\
Papers w/o leaderboards &  4,063 & 604 & 507  \\
%Prompts &  - & - & 119,805 & 35,295 & - &  - \\
Total TDM-triples & 612,709 & 74,475 & 76,936  \\
%Avg. TDMS Tokens & 4.1\tnote{a} & 4.1\tnote{a} & 46.23 & 45.19 &  2.64\tnote{a} & 2.41\tnote{a} \\
Distinct TDM-triples & 62,629 & 8,748 & 8,434  \\
Distinct \textit{Tasks}       & 1,365 & 432 & 339  \\
Distinct \textit{Datasets}    & 4,733 & 1,379 & 1,077  \\
Distinct \textit{Metrics}     & 2,845 & 850 & 825  \\
Avg. no. of TDM per paper & 5.1 & 5.1 & 6.7  \\
Avg. no. of TDMS per paper & 6.9 & 6.3 & 9.0\\
% \botrule
\end{tabular*}
\caption{Full Paper dataset statistics. The ``papers w/o leaderboard'' refers to papers that do not report leaderboard.}
%\begin{tablenotes}
%    \item[a] Avg. number of TDM-triples per paper
%\end{tablenotes}
\label{table:datasetStats}
\end{minipage}
\end{threeparttable}
\end{center}
\end{table*}

\begin{table*}[!ht]
\begin{center}
\begin{threeparttable}
\begin{minipage}{\textwidth}
\begin{tabular*}{\textwidth}{@{\extracolsep{\fill}}l|ccc@{\extracolsep{\fill}}}
% \toprule%
\cmidrule{1-4}%
& \multicolumn{3}{@{}c@{}}{\textbf{Our Corpus}} \\\cmidrule{2-4}%
 & Train &Test-Few-shot & Test Zero-shot \\
\midrule
Papers w/ leaderboards & 7,025 & 903& 573  \\
Papers w/o leaderboards &  3,033 & 444 & 353  \\
%Prompts &  - & - & 119,805 & 35,295 & - &  - \\
Total TDM-triples & 515,203 & 63,041 & 61,474  \\
%Avg. TDMS Tokens & 4.1\tnote{a} & 4.1\tnote{a} & 46.23 & 45.19 &  2.64\tnote{a} & 2.41\tnote{a} \\
Distinct TDM-triples & 56,486 & 8,241 & 7,557  \\
Distinct \textit{Tasks}       & 1,232 & 417 & 323  \\
Distinct \textit{Datasets}    & 4,473 & 1,317 & 1,010  \\
Distinct \textit{Metrics}     & 2,687 & 812 & 778  \\
Avg. no. of TDM per paper & 5.0 & 5.0 & 6.7  \\
Avg. no. of TDMS per paper & 6.8 & 6.2 & 8.9\\
% \botrule
\end{tabular*}
\caption{DOCTEAT dataset statistics. The ``papers w/o leaderboard'' refers to papers that do not report leaderboard.}
%\begin{tablenotes}
%    \item[a] Avg. number of TDM-triples per paper
%\end{tablenotes}
\label{table:datasetStats}
\end{minipage}
\end{threeparttable}
\end{center}
\end{table*}

We used a context length of 2400 and based on GPU availability, a batch size of 2 and gradient\_accumulation\_steps of 4 were used, leading to a final batch size of 8. All experiments were run on five epochs and we used AdafactorSchedule and Adafactor optimizer~\cite{shazeer2018adafactor} with scale\_parameter=True, relative\_step=True, warmup\_init=True, lr=1e-4.

% The evaluations were all done on a dataset made of individual template instructions separately, as reported in table \ref{tab:rouge-50percent} to \ref{tab:f1-all}.

% \section{Corpus}
% \label{sec:corpus}

\end{document}